\documentclass[journal]{IEEEtran}

\usepackage{mathtools}
\usepackage{subfigure}
\usepackage{amsmath}
\usepackage{graphicx}
\usepackage{multirow}
\usepackage{makecell}
\usepackage{float}
\usepackage{cite}
\usepackage{indentfirst}
\usepackage{amsthm}
\usepackage{color}
\usepackage{mathrsfs}
\usepackage{amssymb}
\usepackage{amsbsy}
\usepackage{times}  
\usepackage{url}  
\usepackage{verbatim}
\usepackage{hyperref}
\usepackage{times}  
\usepackage{helvet}  
\usepackage{courier}  
\usepackage{url}  
\usepackage{graphicx}  
\usepackage{mathtools}
\usepackage{mathrsfs}
\usepackage{amsfonts}
\usepackage{multirow}
\usepackage{diagbox}
\usepackage{algorithm}
\usepackage{algorithmic}

\usepackage{comment}

\newtheorem*{theorem*}{Theorem}

\newtheorem*{corollary*}{Corollary}

\newtheorem*{lemma*}{Lemma}

\newtheorem*{proposition*}{Proposition}

\ifCLASSINFOpdf

\else

\fi

\begin{document}

\title{Multi-Hop Convolutions of Weighted Graphs}

\author{Qikui Zhu, Bo Du,~\IEEEmembership{Senior Member,~IEEE}, Pingkun Yan,~\IEEEmembership{Senior Member,~IEEE}%
%
\thanks{Asterisks indicate co-corresponding authors.}%
\thanks{Q.~Zhu is with School of Computer Science,
Wuhan University, Wuhan, China. (e-mail: QikuiZhu@whu.edu.cn).}%
\thanks{*B.~Du is with School of Computer Science and  State Key Lab of Information Engineering on Survey, Mapping and Remote Sensing, Wuhan University, Wuhan, China. (e-mail: remoteking@whu.edu.cn).}
\thanks{*P.~Yan is with the Department of Biomedical Engineering and the Center for Biotechnology and Interdisciplinary Studies at Rensselaer Polytechnic Institute (RPI), Troy, NY, USA 12180. (e-mail: yanp2@rpi.edu).}%
\thanks{The source code of this work is available at the \href{https://github.com/ahukui/Multi-hop-Convolutions-on-Weighted-Graphs}{GitHub repository}}%
}

\maketitle


\begin{abstract}
Graph Convolutional Networks (GCNs) have made significant advances in semi-supervised learning, especially for classification tasks. However, existing GCN based methods have two main drawbacks. First, to increase the receptive field and improve the representation capability of GCNs, larger kernels or deeper network architectures are used, which greatly increases the computational complexity and the number of parameters. Second, methods working on higher order graphs computed directly from adjacency matrices may alter the relationship between graph nodes, particularly for weighted graphs. In addition, the direct construction of higher-order graphs introduces redundant information, which may result in lower network performance. To address the above weaknesses, in this paper, we propose a new method of multi-hop convolutional network on weighted graphs. The proposed method consists of multiple convolutional branches, where each branch extracts node representation from a $k$-hop graph with small kernels. Such design systematically aggregates multi-scale contextual information without adding redundant information. Furthermore, to efficiently combine the extracted information from the multi-hop branches, an adaptive weight computation (AWC) layer is proposed. We demonstrate the superiority of our MultiHop in six publicly available datasets, including three citation network datasets and three medical image datasets. The experimental results show that our proposed MultiHop method achieves the highest classification accuracy and outperforms the state-of-the-art methods. The source code of this work have been released on GitHub (https://github.com/ahukui/Multi-hop-Convolutions-on-Weighted-Graphs).
\end{abstract}


\IEEEpeerreviewmaketitle

\section{Introduction}

\IEEEPARstart{D}{eep} leaning methods exhibit promising performance in many fields\cite{7078921}, such as computer vision\cite{8275511,8435966}, natural language processing (NLP), and medical image computing\cite{7864335}. In addition to the success of deep convolutional neural networks (CNNs) in grid-structured data analysis, there is an increased interest in applying deep learning, especially Graph Convolutional Networks (GCNs), to arbitrarily structured data like social network, knowledge graphs and chemical molecules \cite{duvenaud2015convolutional,monti2017geometric,atwood2016diffusion}.
For instance, Hamilton et~al.~ \cite{hamilton2017inductive} proposed a general inductive framework called GraphSAGE to efficiently generate node embedding by sampling and aggregating features from a node's local neighborhood for node-classification tasks. Velickovic et~al.~\cite{velivckovic2017graph} developed graph attention networks (GATs), leveraging masked self-attentional layers for performing node classification of graph-structured data. Monti et~al.~\cite{monti2017geometric2} proposed a novel matrix completion architecture for recommendation systems by combining a multi-graph CNN and a recurrent neural network.

GCNs have also started to make an impact in the domain of medical imaging. Before that, the conventional medical image based disease prediction or classification solely relies on the information from images and the relationship and similarity between subjects indicated by non-imaging features have been largely ignored. To efficiently exploit the wealth of both imaging and non-imaging information, for example age, health history, etc., for improving the accuracy of disease prediction, efforts of using GCNs have been made. For example, Parisot et~al.~\cite{parisot2017spectral} introduced GCN for population based disease prediction, which treats a patient population as a sparse graph. Vertices of the graph are associated with image-based feature vectors and the edges encode phenotypic information. To analyze the impact and relevance of the neighborhood definitions on the task of disease prediction, Kazi et~al.~\cite{kazi2018multi} incorporated a novel weighting layer into the GCNs, which automatically learns the weight of each meta-data entry with respect to its relevance to the prediction task. Furthermore, inspired by the success of the inception architecture in CNNs, Kazi et~al.~\cite{kazi2019inceptiongcn} proposed a novel InceptionGCN model, which leverages spectral convolutions with different kernel sizes and chooses optimal features to solve the disease prediction problem.

\begin{figure}[t]
  \centering
  \includegraphics[width=\columnwidth]{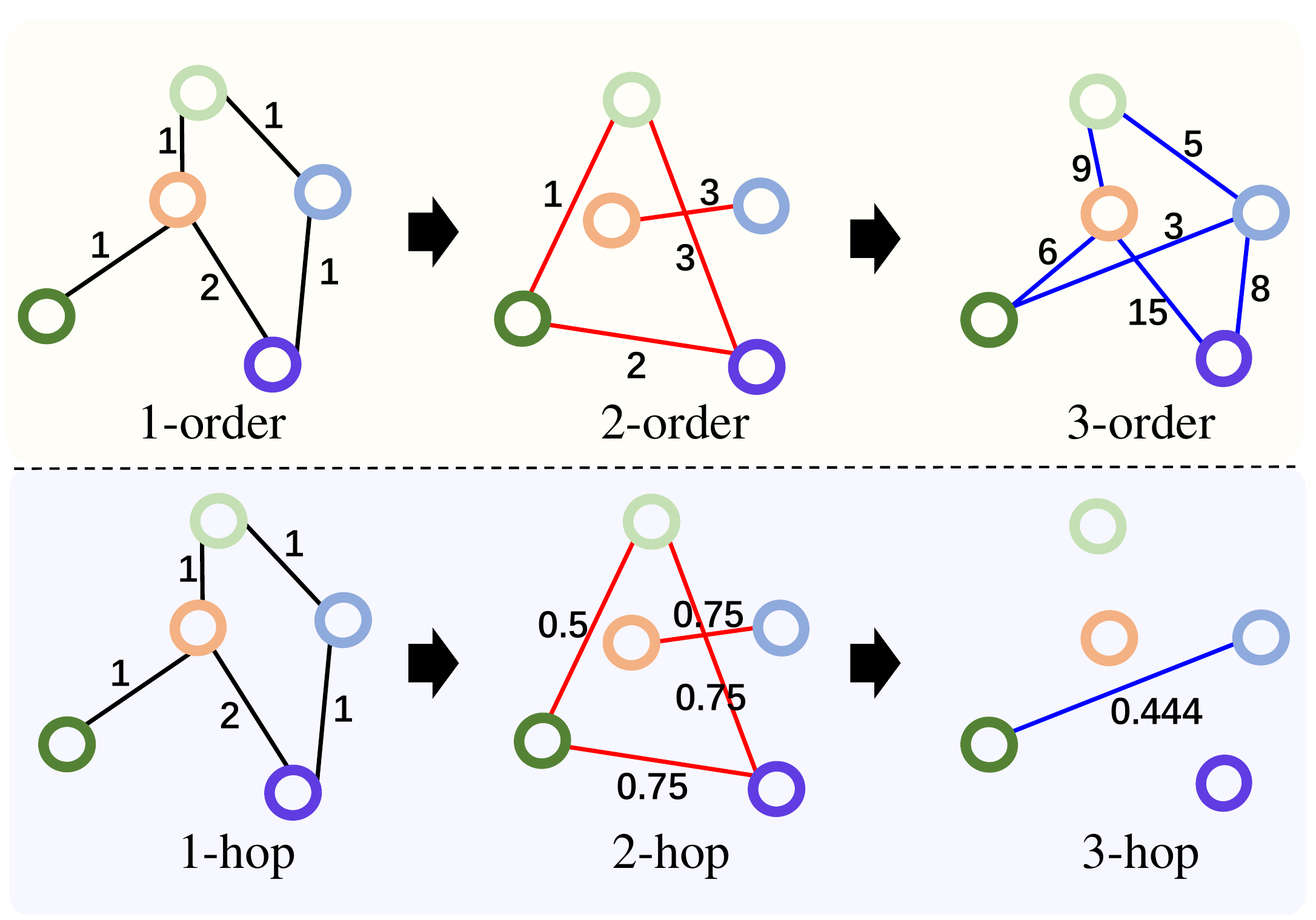}
  \caption{Illustration of the difference between higher order (top) and multiple hop (bottom) graphs. Self-loops are not shown in the figure for simplicity.}\label{OrderHop11}
\end{figure}

Although the above methods have improved the efficiency of GCNs in learning node representation and hence the accuracy of node classification, the structure of the constructed graphs limits the information extraction to be in local neighborhoods, similar to a fixed kernel is used for a CNN. A main reason is because the features of every node can only be learned from its neighbors at a fixed number of hops away.
To overcome this challenge, many researchers try to employ bigger kernel size for GCN filter~\cite{kazi2019inceptiongcn}, or design deeper GCNs~\cite{xu2018representation}, or mix feature representations of neighbors at various distances for constructing special graph structures. For example, Abu-El-Haija et~al.~\cite{abu2019mixhop} proposed MixHop that can learn a general class of neighborhood mixing relationships by repeatedly
mixing feature representations of neighbors at various distances, where nodes receive latent representations from first-degree neighbors and further $N$-degree neighbors at every message passing step. However, utilizing bigger kernel size, increasing the depth of GCNs significantly or employing higher-order graph convolutional network would increase the computational complexity of the convolution operator and the number of parameters. 
For the higher-order graph convolutional networks, such as~\cite{abu2019mixhop}, which construct the higher-order graph by directly multiplying the adjacency matrix many times. Although this operation is easily and effective in higher-order graph constructing, the constructed higher-order graph always bring redundant information.
Besides, this operation also ignored the weight of graph and changed the original relationship between the nodes. For example, as shown in the top row of Figure~\ref{OrderHop11}, for the weighted graph, the operation of multiplying the adjacency matrix reassigned higher weight to those indirect connected nodes, which changes the predefined relationship between nodes. And the connections constructed by higher-order graph (3-order) already exist in the lower-order graph (1-order), which also brings many redundant information. 

To tackle the above-mentioned challenges, in this paper, a novel Multi-hop Graph Convolutional Network (MultiHop) is proposed for classification tasks on graphs datasets, particularly for the weighted graph. The proposed MultiHop consists of multiple graph convolution branches. Each branch captures the node representation from its neighbors with a different hop, for systematically aggregating multi-scale contextual information without adding redundant information.
Different from the higher-order graph constructed by multiplying the adjacency matrix, the higher-order graph used in our model are computed by both the hop and weight between each node and avoid problem of relationship be changed and information redundantly as shown in the bottom row of Figure 1. Compared with utilizing bigger kernel size or designing deeper network architecture, our proposed MultiHop significantly reduces the number of trained parameters and the computational complexity. Specifically, to adaptively fuse the node representations learned from multi-branch, an adaptive weight computation (AWC) layer is designed. The AWC layer works together with multi-branch, which receives the learned node representations and adaptively computes the weight of node representations from different branches. MultiHop then fuses the features together for getting the final node representations. In certain sense, the proposed MultiHop to GCN may be analogous to the successful dilated convolutions\cite{yu2015multi} to CNNs.

In this paper, extensive experiments were performed on two categories of datasets, three publicly disease prediction datasets - ABIDE~\cite{di2014autism}, TADPOLE~\cite{marinescu2018tadpole} and Chest X-rays~\cite{jaeger2014two} and three citation network datasets - Citeseer, Cora and Pubmed \cite{yang2016revisiting}, are used to evaluate our proposed methods. The results corroborate the effectiveness of our proposed MultiHop, which outperforms state-of-the-art methods.

In the rest of the paper, we first give a short overview the spectral graph convolutional networks (GCNs). Then we will present our proposed the Multi-Hop GCN in detail, which is followed by the experimental results and discussions.


\begin{figure*}[t]
  \centering
  \includegraphics[width=\textwidth]{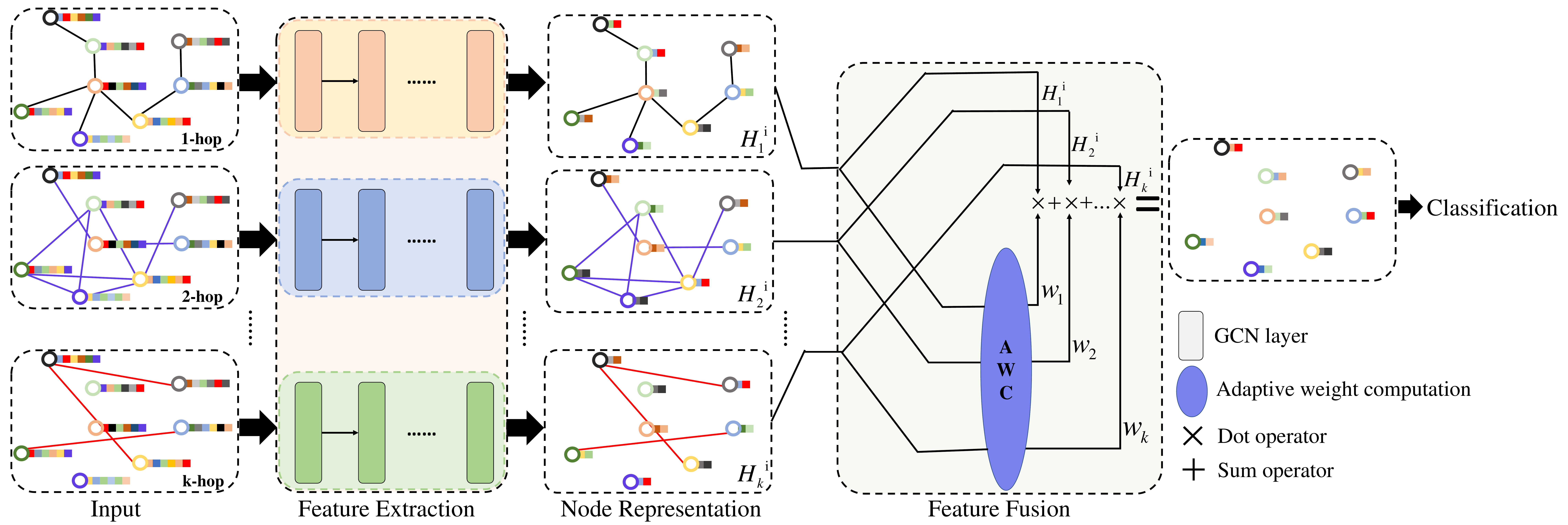}
  \caption{Overview of the proposed Multi-hop Graph Convolutional Network.}\label{ProposedModel}
\end{figure*}

\section{Spectral Graph Convolutional Networks}

In this section, we briefly review the mathematical background of spectral graph convolution network employed in our work. Let $G({\cal V},{\cal E})$ represents an undirected graph with $N$ nodes ${v_i} \in {\cal V}$, edges $({v_i},{v_j}) \in {\cal E}$. And ${\mathop{\rm A}\nolimits}  \in {\mathbb{R}^{N \times N}}$ is the adjacency matrix and ${\mathop{\rm D}\nolimits}  \in {\mathbb{R}^{N \times N}}$ is the degree matrix of $G({\cal V},{\cal E})$, where ${{\mathop{\rm D}\nolimits} _{ii}} = \sum\nolimits_j {{{\mathop{\rm A}\nolimits} _{ij}}}$ . The normalized graph Laplacian matrix is defined as $L = {I_N} - {D^{ - \frac{1}{2}}}{\mathop{\rm A}\nolimits} {D^{ - \frac{1}{2}}}$, which can be factored as $L = U\Lambda {U^T}$, where $U$ is the matrix of eigenvectors ordered by eigenvalues and $\Lambda$ is the diagonal matrix of eigenvalues. Accordingly, for a graph signal $x \in {\mathbb{R}^{N}}$ which is the feature vector for every nodes of a graph. Giving a filter ${g_\theta } = diag(\theta )$ which is a diagonal matrix filled with learnable parameters $\theta \in {\mathbb{R}^{N}}$ in the Fourier domain. The spectral graph convolution operator on $x$ can be defined as:
\begin{align}\label{Eq1}
{g_\theta } * x = U{g_\theta }{U^T}x,
\end{align}
where ${U^T}x$ is the graph Fourier transform of $x$.
Based on the definition of spectral graph convolution, Bruna et~al.~\cite{bruna2013spectral} first proposed the spectral convolution neural network (Spectral CNN on graphs). They define a spectral convolution layer as:
\begin{align}\label{Eq2}
{H^{i + 1}} = \sigma (U{g_\theta }{U^T}{H^{i}}),
\end{align}
where ${H^{i}} \in {\mathbb{R}^{N \times M}}$  and ${H^{i + 1}} \in {\mathbb{R}^{N \times M'}}$  are the input and output of Spectral CNN layer, which also are the feature descriptions (where $N$ denotes the number of nodes and $M$ indicates the dimensionality of node features), $\sigma $  denotes a non-linear activation function, such as the ${\mathop{\rm ReLU}\nolimits} ( \cdot ) = max(0, \cdot )$.

However, since multiplication with the eigenvector matrix $U$ is ${\rm O}({N^2})$ and computing the eigendecomposition of $L$ in the first place might be prohibitively expensive for large graphs, which make Spectral CNN compute expensive. To circumvent this challenge, Defferrard et~al.~\cite{defferrard2016convolutional} proposed ChebNet which defines a filter as Chebyshev polynomials of the diagonal matrix of eigenvalues.

The ${g_\theta }$ can be well-approximated by a truncated expansion in terms of Chebyshev polynomials ${T_k}(\tilde \Lambda )$ up to ${K^{th}}$ order:
\begin{align}\label{Eq3}
{g_\theta }(\Lambda ) \approx \sum\limits_{k = 0}^K {{\theta _k}{T_k}(\tilde \Lambda )},
\end{align}
where $\tilde \Lambda  = \frac{2}{{{\lambda _{\max }}}}\Lambda  - {I_N}$. The Chebyshev polynomials are defined recursively by ${T_k}(x) = 2x{T_{k - 1}}(x) - {T_{k - 2}}(x)$ with ${T_0}(x) = 1$ and ${T_1}(x) = 1$.

Based on this, the convolution of a graph signal $x$  with the defined filter ${g_\theta }$ is
\begin{align}\label{Eq4}
{g_\theta } * x \approx \sum\limits_{k = 0}^K {{\theta _k}{T_k}(\tilde \Lambda )} x.
\end{align}
From this equation, we can note that which can reduce the computation complexity of spectral graph convolution to ${\rm O}(\left| {\cal E} \right|)$ linear in the number of edges.

To further alleviate the problem of overfitting on local neighborhood structures for graphs with very wide node degree distributions, Kipf et~al.~\cite{kipf2016semi} proposed a first-order approximation of ChebNet. They limited the layer-wise convolution operation to $k=1$. And therefore the ChebNet becomes a linear function on the graph Laplacian spectrum.
Furthermore, in this linear formulation of a GCN, they further approximate ${\lambda _{\max }} \approx 2$, Under these approximations, Eq.\ref{Eq4} is simplified as
\begin{align}\label{Eq5}
{g_\theta } * x \approx {\theta _0}x - {\theta _1}{D^{ - \frac{1}{2}}}{\mathop{\rm A}\nolimits} {D^{ - \frac{1}{2}}}x.
\end{align}

To further constrain the number of parameters, address overfitting and minimize the number of operations per layer, such as matrix multiplications, they further assumed $\theta  = {\theta _0} =  - {\theta _1}$. Therefor, Eq.\ref{Eq5} can be written as:
\begin{align}\label{Eq6}
{g_\theta } * x \approx \theta ({I_N} + {D^{ - \frac{1}{2}}}{\mathop{\rm A}\nolimits} {D^{ - \frac{1}{2}}})x.
\end{align}
In order to avoid numerical instabilities and exploding or vanishing gradients bring by stacking multiple layers, ${I_N} + {D^{ - \frac{1}{2}}}{\mathop{\rm A}\nolimits} {D^{ - \frac{1}{2}}}$ be replaced by ${\tilde D^{ - \frac{1}{2}}}\tilde A{\tilde D^{ - \frac{1}{2}}}$, with $\tilde A = {I_N} + {\mathop{\rm A}\nolimits}$ and ${\tilde D_{ii}} = \sum\nolimits_j {{{\tilde A}_{ij}}}$. 

Accordingly, for a graph ${X} \in { \mathbb{R}^{N \times C}}$  where $N$ denotes the number of nodes and $C$ indicates the dimension of node features, the definition of graph convolution layer is:
\begin{align}\label{Eq7}
{H^{i + 1}} = {\tilde D^{ - \frac{1}{2}}}\tilde A{\tilde D^{ - \frac{1}{2}}}{H^{i}}\Theta,
\end{align}
where $\Theta  \in { \mathbb{R}^ {C \times F}}$ is a matrix of filter parameters, ${H^{i+1}} \in {\mathbb{R}^{N \times F}}$ is the convoluted matrix, ${H^{0}} = X$ and $F$ denotes the number of filters.


\begin{figure*}
  \centering
  \includegraphics[width=\textwidth]{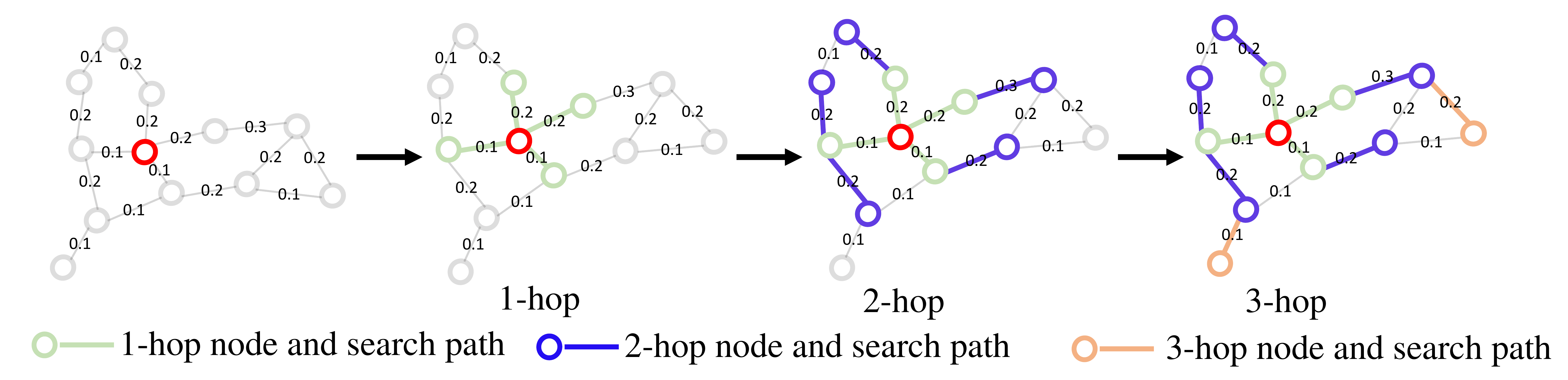}
  \caption{The process of searching various hop neighbors. Numbers on the graph edges are the weights.}\label{HopsSearch}
\end{figure*}

\section{Multi-Hop GCN}

In this section, we first give an overview of the our MultiHop and then discuss each component in detail.

\subsection{Overall Framework}
 The objective of our MultiHop is to learn a classifier, which can classify each node of a graph $G({\cal V},{\cal E})$ into different categories. Figure~\ref{ProposedModel} gives an overview of the proposed MultiHop. As shown in Figure~\ref{ProposedModel}, our proposed MultiHop consists of two major components: \textbf{feature extraction} and \textbf{feature fusion}. The \textbf{feature extraction} component aims to learn feature representations of each node within its neighborhood defined by various hops. Multi-branch graph convolution layers are designed to achieve the feature extraction. The \textbf{feature fusion} component adaptively computes the weights of learned feature representations by our specifically designed AWC layer. The representation of each node from different branches are then aggregated together using the computed weights to obtain the final node representation for classification. Details of the proposed method are presented in the following sections.

\subsection{Affinity Graph Construction}

The construction of an affinity graph $G({\cal V},{\cal E})$ is crucial to accurately model the interactions between the subjects. A graph $G$ consists of $\left| {\cal V} \right|{\rm{ = }}N$ nodes connected by the weighted edges ${\cal E} \in { \mathbb{R}^{N \times N}}$. Mathematically,  ${{\cal E}}$  can be defined as
\begin{align}\label{Eq10}
{\cal E} = {W} \circ A,
\end{align}
where ${W} \in { \mathbb{R}^{N \times N}}$ is the weight matrix, $A \in { \mathbb{R}^{N \times N}}$ is adjacency matrix, and  $\circ$ denotes the Hadamard product. 

For medical image datasets with no prior defined graphs, to efficiently exploit the wealth of both  imaging  and  non-imaging  information, both information sources are used to define a weighted graph $G$ in our work. The patients are set as nodes $\cal V$ with their corresponding node feature vectors extracted from image data. The graph's adjacency matrix $A$ is computed by the non-imaging measures ${M} \in R^n$ (e.g. gender, age or other meta-information) as
\begin{align}\label{Eq11}
A({i},{j}) = \sum\limits_{t = 1}^n {\gamma} ({M_t}(i),{M_t}(j)),
\end{align}
where
\begin{align}\label{Eq8}
{\gamma}({M_t}(i),{M_t}(j)) = \left\{ {\begin{array}{*{20}{c}}
1,&{if\left| {{M_t}(i) - {M_t}(j)} \right| < \beta }\\
0,&\mathrm{otherwise},
\end{array}} \right.
\end{align}
and ${M_t}(i)$ and ${M_t}(j)$ are the value of the $t$ th non-imaging measurement for nodes ${i}$ and ${j}$, respectively. $\beta$ is a threshold for the element similarity.

The weight matrix ${W}$ is computed by the similarities between the feature vectors extracted from imaging data. \emph{i.e.},
\begin{align}\label{Eq9}
W({i},{j}) = \exp \left( - \frac{{{{\left[ {\rho ({i},{j})} \right]}^2}}}{{2{\sigma ^2}}} \right),
\end{align}
where $\rho$ is the correlation distance between nodes $i$ and $j$, and $\sigma$ is the width of the used convolution kernel. 


For those datasets with pre-defined adjacency matrix $A \in { \mathbb{R}^{N \times N}}$, such as citation network datasets
including Citeseer, Cora and Pubmed, we just need to compute the weight matrix ${W} \in { \mathbb{R}^{N \times N}}$ based on the similarities between node features.

By changing the connection of affinity graph ${\cal E}$, we can construct various hop graphs ${{\cal E}_k}$. The details are described in Figure~\ref{HopsSearch} and Algorithm~\ref{algA}.
First, for each node $i$ in graph ${\cal E}$ we search the shortest path from $i$ to its $k$ hop distance away node $i_k$ by depth first search algorithm (DFS). Then, we statistic the weight of each path and select the path with the maximum weight as the new wight for ${{\cal E}_k(i,i_k)}$. To avoid change the relationship between nodes, we use the value of hop $k$ to assign a small value for the graph with huge hop. Figure~\ref{OrderHop11} is an example of various hop graphs ${{\cal E}_k}$ produced by our proposed method. 

\begin{algorithm}
\caption{$k$-hop graph construction}
\label{algA}
\begin{algorithmic}
\STATE {Input $G({\cal V},{\cal E})$, $k$} 
\STATE {${{\cal E}_k}$ = zeros\_like (${\cal E}$)}
\FOR{$v$ in ${\cal V}$}
\STATE \textbf{PathSet} = \textbf{DFS}$({\cal E}, v, k)$ 
\FOR{ $path$ in \textbf{PathSet}}
\STATE $w = \frac{1}{{{k^2}}}[{\cal E}(v,path[1]) + {\cal E}(path[1],path[2]) + ... + {\cal E}(path[k-1],path[k])]$
\IF{$w > {{\cal E}_k}(v,path[k])$}
\STATE ${{\cal E}_k}(v,path[k]) = w $ 
\ENDIF{}
\ENDFOR{} 
\ENDFOR{}
\STATE {Return ${{\cal E}_k}$} 
\end{algorithmic}
\end{algorithm}

\begin{figure*}
	\centering
	\includegraphics[width=.8\textwidth]{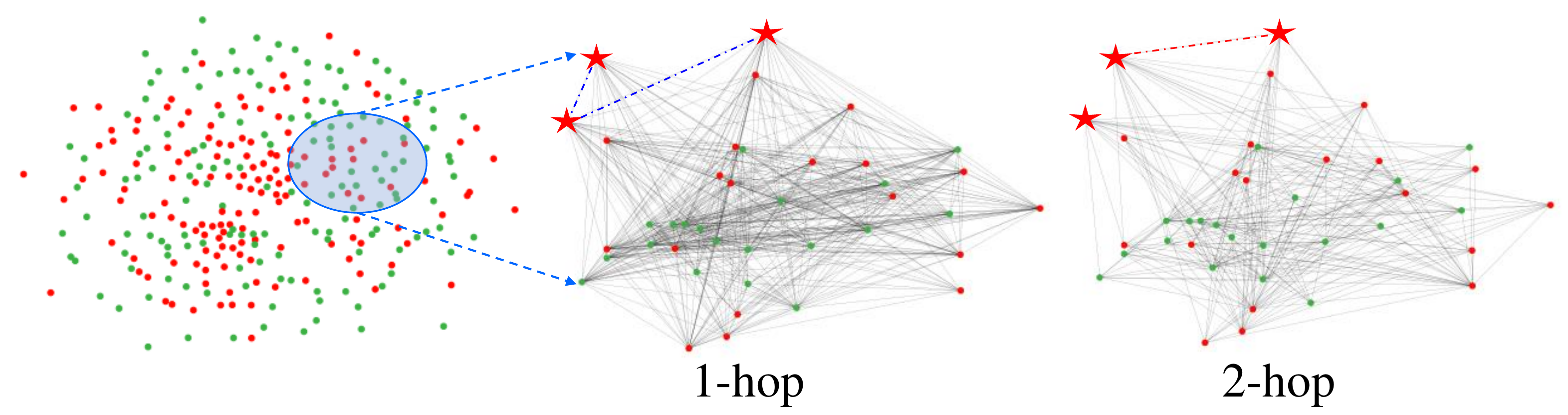}
	\caption{Affinity graph for part of CXR dataset. Different colors represent different categories. The various hop graphs change the connection manner between each node.  As shown in star nodes, the blue doted lines exist in 1-hop graph and disappeared in 2-hop graph, new red doted line be created in 2-hop graph.}
	\label{CXRhops}
\end{figure*}

\subsection{Multi-branch GCN}

GCNs are commonly implemented by using many graph convolutional layers in a sequence manner with a constant kernel size~\cite{parisot2017spectral}. For a single graph convolutional layer in GCN, it only can learn the features of each node from its neighbors within a fixed number of hops. Applying GCN with fewer layers on the whole graph might not capture the semantic and comparable feature from the whole graph. In order to enlarge the receptive field, the network has to be deep or employs big kernels, for example the work in~\cite{kazi2019inceptiongcn}. However, such design of GCNs may significantly increase the computational complexity of the convolution operation and the number of parameters, which may lead to overfitting and decrease the network performance.

To overcome the above challenges and enlarge the learning field of each node, in this paper, we utilize various hop graphs with weighted edges to directly represent the relationship between nodes. Inside the various hop graphs, each node can directly connect with farther neighbors, which enables each node can directly learn the node representation from its neighbors with greater distances by the spectral convolutions with a small kernel. To learn the node representations from above hop graphs, we employ multiple branches in our MultiHop with each branch aggregating the representations of node from unique hop graph respectively as shown in Figure~\ref{ProposedModel}. Compared with those models utilizing larger kernel size or deeper architecture for increasing the receptive fields of convolutional filters, our proposed model is more effective and directly in learning the representations of node from large receptive field with fewer training parameters.

\subsection{Adaptive Weight Computation}

To adaptively fuse the node representations learned from the multiple branches, inspired by the work of Velickovic et~al.~\cite{velivckovic2017graph}, we propose a novel adaptive weight computation (AWC) layer. 
The input of AWC layer is a set of node features $\{ {H_1}^i,{H_2}^i,...,{H_k}^i\}$ from the $i$th graph convolution layer, ${H_k}^i \in {\mathbb{R}^{N \times C}}$ is the nodes represents learned by the $k$th branch, $N$ is the number of nodes and $C$ is the number of classes. The AWC layer computes the weight coefficients $\{ \boldsymbol{w_1},\boldsymbol{w_2},...,\boldsymbol{w_k}\}$ of node features, $\boldsymbol{w_i} \in {\mathbb{R}^{N}}$, and outputs a fusing node representation ${H}^i$. The above process is accomplished through a single-layer feed-forward neural network, parameterized by a shared weight vector $ {\boldsymbol{\alpha }} \in {\mathbb{R}^{C}}$.
Formally, the weights are computed by
\begin{align}\label{Eq13}
\boldsymbol{w_j} = \frac{{\exp (\tanh ({H_j}^i\boldsymbol{\alpha} ))}}{{\sum\limits_{t = 1}^k {\exp (\tanh ({H_t}^i\boldsymbol{\alpha} ))} }},
\end{align}
{and the final feature is}
\begin{align}\label{Eq14}
{H}^{i} = \sum\limits_{t = 1}^k \boldsymbol{w_t}{H_t}^i.
\end{align}

Through the above AWC layer, our MultiHop GCN can adaptively aggregate the representations of each node learned from multiple branches. It enables the capture of both local and global structural information from 1 to $k$-hop neighbors.

\section{Experiments and Results}

\subsection{Datasets and Implementation Details}

In our work, six publicly available datasets belonging to two categories (medical image datasets and citation network datasets) are used for evaluating the proposed methods. The data description and implementation details are presented as follows.

\subsubsection{Dataset 1 -- ABIDE:} The Autism Brain Imaging Data Exchange (ABIDE) database \cite{di2014autism} is a collaboration of different international imaging sites that have aggregated and are openly sharing neuroimaging data from 539 individuals suffering from ASD and 573 age-matched typical controls (TC; 7–64 years). These 1112 datasets are composed of structural and resting state functional magnetic resonance imaging (R-fMRI) datasets along with an extensive array of phenotypic information for Autism spectrum disorders (ASD) studies.
For fair comparison, we choose the same set of 871 subjects used by \cite{parisot2017spectral,kazi2019inceptiongcn}, which are divided into normal (468) and ASD diseased (403) subjects. We also follow the same pre-processing step as performed in \cite{parisot2017spectral,kazi2019inceptiongcn}. The pre-processing consists of skull striping, slice timing correction, motion correction, global mean intensity normalisation, nuisance signal regression, band-pass filtering (0.01-0.1Hz) and registration of the functional MRI images to MNI152 standard anatomical space.
We follow the same graph construction step as in the baseline method~\cite{parisot2017spectral}. Two non-imaging measures, gender and acquisition site, are used for constructing the graph. The 200 most discriminating features from each subject are selected by a ridge classifier as subject's features. For the  element similarity threshold $\beta$, we set $\beta=0$ for the two non-imaging elements. A MultiHop network with two branches is constructed for this datasets. Each branch has two graph convolutional layers. The filter number of the two layers are 16 and 2, respectively.

\subsubsection{Dataset 2 -- TADPOLE:} This dataset \cite{marinescu2018tadpole} is derived from the Alzheimer’s Disease Neuroimaging Initiative (ADNI) database (adni.lonu.usc.edu), which consists of 557 patients with 354 multi-modal features per patient. There are total three classes existence, Cognitively Normal (CN), Mild Cognitive Impairments (MCI) or Alzheimer’s Disease (AD). Our goal on the TADPOLE database is to predict the class of each patient. To provide a fair evaluation for TADPOLE, in the experiment, we follow the same pre-processing step as performed in the work of \cite{kazi2019inceptiongcn}. Four demographics are selected (age, gender, APOE status and FDG PET) for graph construction. We also set threshold $\beta=2$ for age and $\beta = 0$ for the rest of the three demographics. The features are extracted from MR and PET imaging, cognitive tests, CSF and clinical assessments. A MultiHop network with four branches is designed. Each branch has two graph convolution layers with 16 and 3 filters, respectively.

\subsubsection{Dataset 3 -- CXR:} 
In this CXR dataset~\cite{jaeger2014two}, the Chest X-rays are from out-patient clinics and captured as part of the daily routine using Philips DR Digital Diagnose systems. The dataset consists of x-ray images from 662 subjects, which are divided into 326 normal and 336 abnormal x-rays showing various manifestations of tuberculosis. Each subject contains two non-imaging elements (gender, age). In our experiment, a pre-trained ResNet~\cite{he2016deep} is employed to extract the feature representations (each feature vector with length of 2048) of all CXR images. The gender and age  are used for graph construction, and we set the element similarity threshold $\beta=5$ for age and $\beta = 0$ for gender. Figure~\ref{CXRhops} shows the affinity graph of part of the CXR dataset. The structure of the network is the same as the one used for the ABIDE dataset.

\begin{table}[t]
\caption{Summary statistics of the citation network datasets 4--6. C indicates the number of categories. LR is the label rate, which is the ratio of the number of labeled nodes used for training over the total number of nodes in each dataset.}
\centering
	\begin{tabular}{l|r|r|c|r|c}
		\hline
		\textbf{Dataset}  & \# Nodes & \# Edges & C & \# Features & LR \\ \hline
		Citeseer & 3,327  & 4,732  & 6       & 3,703     & 0.036      \\ \hline
		Cora     & 2,708  & 5,429  & 7       & 1,433     & 0.052      \\ \hline
		Pubmed   & 19,717 & 44,338 & 3       & 500      & 0.003      \\ \hline
	\end{tabular}
	\label{Tabel_citation}
\end{table}

\begin{table*}[t]
\caption{Mean accuracies of the stratified 10-fold cross validation on the medical image datasets. ``Mixed'' indicates that the affinity graph is constructed by averaging all the graphs built with different measures. ``Best'' denotes that the best performance obtained under the affinity graph constructed with a particular non-imaging measure.}
  \centering
\begin{tabular}{l|c|l|l|c}
\hline
\multicolumn{2}{c|}{\textbf{Dataset}}    & GCN~\cite{parisot2017spectral} & InceptionGCN~\cite{kazi2019inceptiongcn} & MutiHop (ours)                                  \\ \hline
\multirow{2}{*}{\textbf{ABIDE}}   & Mixed & 67.86 $\pm$ 5.93                                     & 66.70 $\pm$ 6.27                                                 & \multirow{2}{*}{\textbf{70.04 $\pm$ 4.52}} \\ \cline{2-4}
                         & Best  & 67.86 $\pm$ 5.93 (Mixed)                                     & 67.97 $\pm$ 4.43 (Site)                                        &                                                             \\ \hline
\multirow{2}{*}{\textbf{TADPOLE}} & Mixed & 82.04 $\pm$ 5.71                                       & 84.35 $\pm$ 6.97                                               & \multirow{2}{*}{\textbf{91.82 $\pm$ 3.07}} \\ \cline{2-4}
                         & Best  & 84.59 $\pm$ 4.82 (Gender)                            & 88.53 $\pm$ 3.27 (Age)                                         &                                                             \\ \hline
\multirow{2}{*}{\textbf{CXR}}     & Mixed & 84.39 $\pm$ 4.55                                       & 86.52 $\pm$ 5.14                                               & \multirow{2}{*}{\textbf{87.12 $\pm$ 5.13}} \\ \cline{2-4}
                         & Best  & 85.76 $\pm$ 3.26(Age)                                & 86.52 $\pm$ 5.14 (Mixed)                                       &                                                             \\ \hline
\end{tabular}
\label{Tabel_medical}
\end{table*}

\begin{table*}
\caption{Results of node classification on the citation datasets from~\cite{yang2016revisiting}}
  \centering
\begin{tabular}{l|c|c|c}
\hline
\textbf{Model}                               & \textbf{Citeseer}              & \textbf{Cora}                  & \textbf{Pubmed}                \\ \hline
ManiReg~\cite{belkin2006manifold}                             & 60.1                  & 59.5                  & 70.7                  \\
SemiEmb~\cite{weston2012deep}                             & 59.6                  & 59.0                  & 71.1                  \\
LP~\cite{zhu2003semi}                                  & 45.3                  & 68.0                  & 63.0                  \\
DeepWalk~\cite{perozzi2014deepwalk}                            & 43.2                  & 67.2                  & 65.3                  \\
ICA~\cite{lu2003link}                                 & 69.1                  & 75.1                  & 73.9                  \\
Planetoid~\cite{yang2016revisiting}                       & 64.7                  & 75.7                  & 77.2                  \\
Vanilla GCN~\cite{kipf2016semi}                         & 70.3                  & 81.5                  & 79.0                  \\
MixHop~\cite{abu2019mixhop}                            & 71.4                  & 81.9                  & \textbf{80.8}                  \\ \hline
MutiHop (ours) & \textbf{71.5 $\pm$ 0.82} & \textbf{82.4 $\pm$ 0.57} & 79.4$\pm$ 0.26 \\ \hline
\end{tabular}
\label{Tabel_cite}
\end{table*}

\begin{table*}[t]
  \caption{Mean accuracies of the stratified 10-fold cross validation on the TADPOLE dataset with varying number of branches.}
  \centering
\begin{tabular}{c|c|c|c|c|c|c}
\hline
 \backslashbox{Fusion}{\# of Branches}& 2 & P-Value  & 3 & P-Value  & 4 & P-Value \\ \hline
 max-pooling & 88.55 $\pm$ 3.64 & 0.0064 & 87.09 $\pm$ 5.78 & 0.0027 & 87.09 $\pm$ 5.23 & 0.0188 \\ \hline 
 sum & 89.09 $\pm$ 4.95& 0.0809 & 88.18 $\pm$ 4.00& 0.0015 & 89.45 $\pm$ 3.43 & 0.0225\\ \hline
 AWC & \textbf{91.09 $\pm$ 3.09} & --- & \textbf{91.27 $\pm$ 3.23}& --- & \textbf{91.82 $\pm$ 3.07} & --- \\ \hline
\end{tabular}
\label{Tabel_TADPOLE}
\end{table*}

\begin{table}
  \caption{Mean accuracies of the stratified 10-fold cross validation on the ABIDE and CXR datasets.}
  \centering
 	\fontsize{8.5}{10}\selectfont
\begin{tabular}{c|c|c|c|c}
\hline
\multirow{2}{*}{Fusion} & \multicolumn{2}{c|}{ABIDE}                           & \multicolumn{2}{c}{CXR}                             \\ \cline{2-5} 
                        & Acc                                        & P-Value & Acc                                        & P-Value \\ \hline
max-pooling            & 66.36 $\pm$ 4.24                           & 0.0217  & 84.70 $\pm$ 4.81                           & 0.0016  \\ \hline
sum                     & 67.28 $\pm$ 4.01                           & 0.0423  & 86.21 $\pm$ 4.76                           & 0.0406  \\ \hline
AWC                     & \textbf{70.04 $\pm$ 4.52} & ---     & \textbf{87.12 $\pm$ 5.13} & ---     \\ \hline
\end{tabular}
\label{Tabel_ABIDE_CXR}
\end{table}

\subsubsection{Datasets 4--6  -- Citation Network Datasets:} 
We also evaluate the effectiveness of our proposed model following the experimental setup in \cite{yang2016revisiting} using three citation network datasets - Citeseer, Cora and Pubmed, which belong to the second category. The nodes with sparse bag-of-words feature vectors represent document and the edges denote citation links between documents. Table~\ref{Tabel_citation} summarizes the dataset statistics.

Our experiments on the citation network datasets follow the setup presented in~\cite{yang2016revisiting}. Differently, to prove the effectiveness of our MultiHop in weighted graph, in our experiment, we use the similarity between the feature vectors at every node to weight the adjacency matrix of graph. Since the feature vectors of Citeseer and Core are binary and sparse, we employed $L1$ distance to compute the weight matrix. The network architecture consists of three branches. Each branch has two graph convolution layers. The number of filter in the first layer is 16. The number of filter in the second layer equals to the number of categories. 
For the Pubmed dataset, each publication is described by a TF/IDF weighted word vector from a dictionary consisting of 500 unique words. The correlation distance is employed to evaluate the similarity. The network architecture has two  branches and each branch has two graph convolution layers with 8 and 3 filters, respectively. 

\subsection{Implementation Details}
The proposed method is implemented using the open-source deep learning library Keras \cite{chollet2015keras}. Each model is trained end-to-end with Adam optimizer. $L_2$ regularization with weight of 0.0005 and dropout in input and hidden layers with rate of 0.5 are also be used. The activation function is ELU~\cite{clevert2015fast}. For medical image datasets, early stopping is utilized, and the spectral graph convolutional layer employed in our model is approximated by the order $K = 3$ Chebyshev polynomials~\cite{defferrard2016convolutional}. For a graph signal $x$, convolution with a filter ${g_\theta }$ approximated by Chebyshev polynomials is defined as
\begin{align}\label{Eq4}
{g_\theta } * x \approx \sum\limits_{k = 0}^K {{\theta _k}{T_k}(\tilde \Lambda )} x,
\end{align}
where filter ${g_\theta } = diag(\theta )$ is a diagonal matrix filled with learnable parameters $\theta \in {\mathbb{R}^{N}}$ in the Fourier domain, $\tilde \Lambda  = \frac{2}{{{\lambda _{\max }}}}\Lambda  - {I_N}$, and $\Lambda$ is the diagonal matrix of eigenvalues for the normalized graph Laplacian matrix ${I_N} - {D^{ - \frac{1}{2}}}{\mathop{\rm A}\nolimits} {D^{ - \frac{1}{2}}}$. The Chebyshev polynomials are defined recursively by ${T_k}(x) = 2x{T_{k - 1}}(x) - {T_{k - 2}}(x)$ with ${T_0}(x) = 1$ and ${T_1}(x) = 1$.

For the citation network datasets, the graph convolutional layers used in our model are approximated by first-order Chebyshev polynomial~\cite{kipf2016semi}, which is defined as 
\begin{align}\label{Eq7}
{H^{i + 1}} = {\tilde D^{ - \frac{1}{2}}}\tilde A{\tilde D^{ - \frac{1}{2}}}{H^{i}}\Theta
\end{align}
where $\Theta  \in { \mathbb{R}^ {C \times F}}$ is a matrix of filter parameters, ${H^{i+1}} \in {\mathbb{R}^{N \times F}}$  and ${H^{i}}$ are the output and input of layer and $F$ denotes the number of filters, $\tilde A = {I_N} + {\mathop{\rm A}\nolimits}$ and ${\tilde D_{ii}} = \sum\nolimits_j {{{\tilde A}_{ij}}}$. 

During training, we first train our model about 2000 epochs with an initial learning rate of 0.005, and then train the model about 1000 epochs with the learning rate of 0.001. Since we note that the citation datasets are sensitive to the initialization, we ran all the models 100 times. After ranking the networks by using validation accuracy, we finally report the average test accuracy of the top 50 runs.

\begin{figure*}
  \centering
  \includegraphics[width=\textwidth]{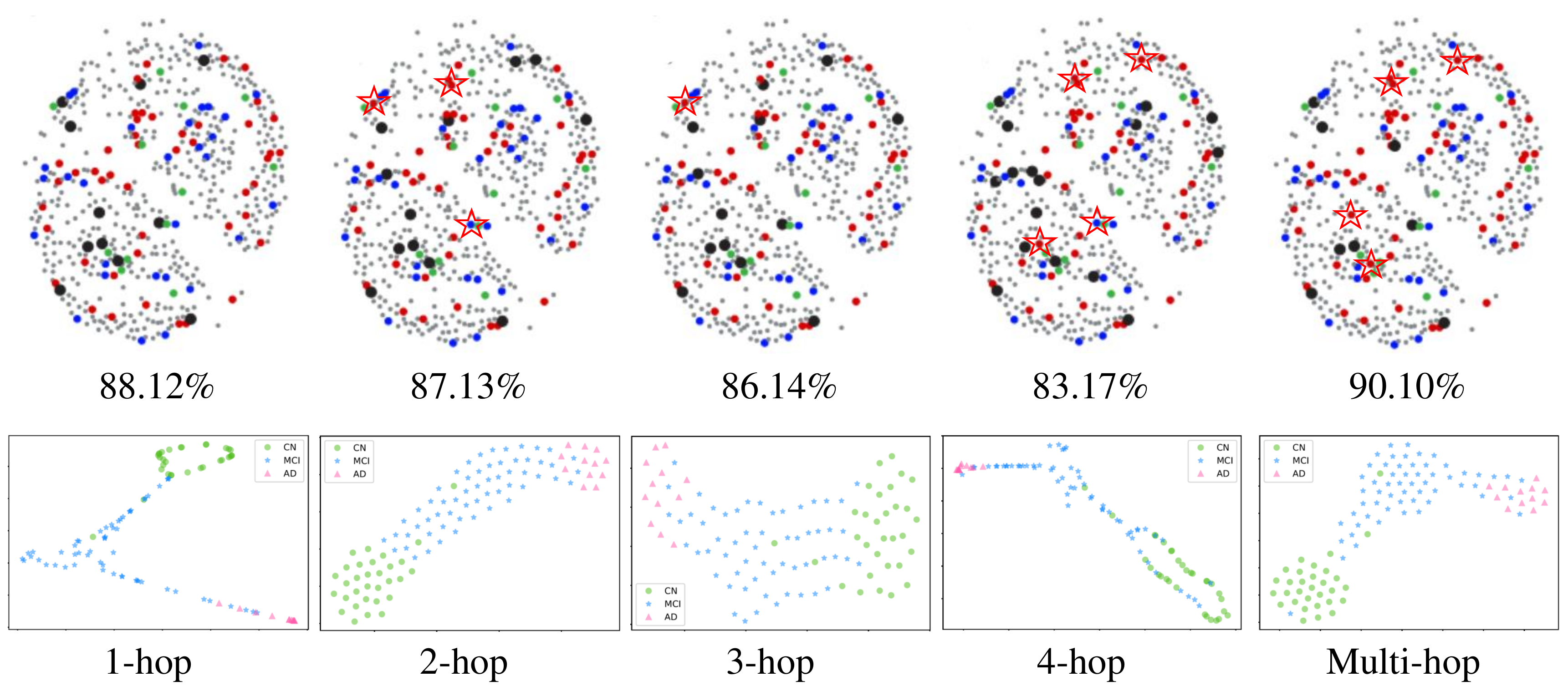}
  \caption{\textbf{Top row}: Classification results of graphs with various hops. Gray nodes represent training data. Green, red and blue nodes are test data of different  categories. The large black nodes are wrongly predicted nodes. The red stars label those nodes, which are incorrectly classified on the 1-hop graph, but correctly predicted in other graphs. The associated classification accuracy is given below the graphs. \textbf{Bottom row}: t-SNE visualization of the obtained features using each hop graph.}\label{Ab}
\end{figure*}
\subsection{Experimental Results}

In this section, we first compare our proposed MultiHop method against several state-of-the-art GCN based methods. Then we evaluate the effectiveness of multi-branch strategy and AWC layer through ablation studies. After that, we further visualize the various hop graphs to investigate feature learning.

\textbf{Comparison with state-of-the-art GCN methods:} To evaluate the performance of our proposed MultiHop, we compare the results against several state-of-the-art methods based on GCN. Table~\ref{Tabel_medical} and Table~\ref{Tabel_cite} list the  performances of our proposed MultiHop and other state-of-the-art GCN methods under two categories of available datasets. 
From Table~\ref{Tabel_medical}, we can note that our MultiHop outperforms other state-of-the-art GCN methods and performed the best on the medical image datasets, which indicates that our MultiHop are generalized and robust to the medical image datasets. It's notable that we used mixed graph for every medical image datasets for creating various hop graphs. 
Particularly, under the same mixed graph, our model outperforms GCN~\cite{parisot2017spectral}, InceptionGCN by an average margin of 2.18\%, 3.34\% for ABIDE dataset, 9.78\%, 7.47\% for TADPOLE dataset, 2.73\%, 0.6\% for CXR dataset, which proved the effectiveness of our MultiHop. The node classification accuracy on citation network datasets are shown in Table~\ref{Tabel_cite}. Compared with other GCN based methods, the proposed MultiHop obtained significantly better performance on both Citeseer and Cora datasets, and very similar performance to MixHop on Pubmed dataset.

\textbf{Effectiveness of multi-branch and AWC layer:} In our experiments, we compare the performances of our MultiHop under different number of branches and three different types of fusing strategies: sum, max-pooling and our proposed AWC layer. The sum operation directly adds all of features learned from multi-branch together, and the max-pooling operation selects the max one from all of features learned by multi-branch. Table~\ref{Tabel_TADPOLE} shows the results of our MultiHop with different number of branch and fusing strategies on TADPOLE datasets, respectively. As it can be seen from Table~\ref{Tabel_TADPOLE}, 
compared with sum and max-pooling, our MultiHop with AWC layer can obtain better performance with the number of branch increasing, which proves that the effectiveness of AWC layer in adaptive computing weight.  Apart from the evaluation criterion of accuracy, we also perform statistical comparison of the results using paired $t$-test with a confidence interval of 0.95 for showing the effectiveness of our proposed AWC layer. Our MultiHop with AWC layer is compared to model with sum and max-pooling layer for statistical significance, and all the $p$ values are also given in corresponding Table~\ref{Tabel_TADPOLE}. It can be seen that our MultiHop with AWC layer significantly outperforms the other fusing strategy  with $p<0.05$, which clearly  demonstrates that AWC layer is effective in improving the performance of classification. We also compare three different types of fusing strategies on ABIDE and CXR datasets, the classification results are listed in Table~\ref{Tabel_ABIDE_CXR}. Compared with sum and max-pooling, the AWC layer significantly improves the performance of model, which further prove that AWC layer are effective in improving the performance of model. 

\textbf{Contribution of various hop graphs:} To demonstrate the information from various hop graphs can improve the performance of model, we use individual graph with different hops to train the  MultiHop which only has one branch. The experiments are carried on TADPOLE dataset, 20\% data be random selected for test and the rest are used for training. The experiment results from four individual graphs are show in Figure~\ref{Ab}. From top row of Figure~\ref{Ab}, we can note that parts of wrongly predicted nodes on 1-hop graph can be reassigned a true label in other graphs, which indicates that the features from the neighbors at different distances are meaningful. We also employ t-SNE \cite{maaten2008visualizing} to visualize the distribution of the node representation learned by each hop graph as shown in bottom row of Figure~\ref{Ab}. We can note that the feature learned by our MultiHop is distinguished, which prove that our MultiHop can make those hop graphs complement each other for improving the accuracy of classification. 

\section{Conclusion}
In this paper, we analyzed the disadvantages of existing GCNs methods on semi-supervised learning. For the methods used bigger kernels or deeper network architectures, which increase the computational complexity and the number of parameters, and for the methods employed higher order graph convolutional networks, which ignore the weight of graph and change the original relationship between the nodes and always bring redundant information. 
To address the two problems, in this paper, we presented a novel Multi-hop Graph Convolutional Network (MultiHop) for classification tasks on graphs datasets, particularly for the weighted graph. The proposed MultiHop can capture the node representation from various order graphs by multi-branch. To aggregate the lower-order and higher-order neighborhood information in an adaptive manner, an adaptive weight computing layer also be proposed. Extensive experiments were conducted to demonstrate the effectiveness of the proposed method. Our proposed MultiHop achieves superior results compared with other state-of-the-art methods.

\bibliographystyle{IEEEtran}
\bibliography{newref}





\end{document}